\documentclass{article}
\usepackage{spconf,amsmath,graphicx}
\usepackage{fixmath}
\usepackage{hyperref}
\usepackage{amsmath,amssymb}
\DeclareMathOperator{\E}{\mathbb{E}}
\let\OLDthebibliography\thebibliography
\renewcommand\thebibliography[1]{
  \OLDthebibliography{#1}
  \setlength{\parskip}{0pt}
  \setlength{\itemsep}{1pt plus 1ex}
}

\begin{document}
\title{Improving Document Binarization via Adversarial Noise-Texture Augmentation}
%
\name{Ankan Kumar Bhunia\textsuperscript{1}, Ayan Kumar Bhunia\textsuperscript{2}\sthanks{Corresponding Author}, Aneeshan Sain\textsuperscript{3},  Partha Pratim Roy\textsuperscript{4}}
\address{\textsuperscript{1}Jadavpur University, India \hspace{0.1cm} \textsuperscript{2}Nanyang Technological University, Singapore\\  \textsuperscript{3}Cognizant Technology Solutions, India \hspace{0.1cm} \textsuperscript{4}Indian Institute of Technology Roorkee, India \\
{\tt\small \textsuperscript{2}ayanbhunia@ntu.edu.sg }
}

%
%
%

%
\maketitle
\begin{abstract}
Binarization of degraded document images is an elementary step in most of the problems in document image analysis domain. The paper re-visits the binarization problem by introducing an adversarial learning approach. We construct a Texture Augmentation Network that transfers the texture element of a degraded reference document image to a clean binary image. In this way, the network creates multiple versions of the same textual content with various noisy textures, thus enlarging the available document binarization datasets. At last, the newly generated images are passed through a Binarization network to get back the clean version. By jointly training the two networks we can increase the adversarial robustness of our system. Also, it is noteworthy that our model can learn from unpaired data. Experimental results suggest that the proposed method\footnote{The full source code of the proposed system is publicly available at \url{https://github.com/ankanbhunia/AdverseBiNet}} achieves superior performance over widely used DIBCO datasets. 
\end{abstract}

\section{Introduction}
\label{sec:intro}
Document image binarization is a fundamental problem in the field of Document analysis. 
Although binarization seems to be quite easy for images of uniform distribution, it can be challenging under real-world scenarios where the document images suffer from various degradations due to aging effect, inadequate maintenance, ink stains, faded ink, bleed-through background, wrinkles, warping  effect, non-uniform variation of intensity and lighting conditions during document scanning. 
\setlength\textfloatsep{5mm}
\begin{figure}
\label{pic1}
  \includegraphics[width=1\linewidth]{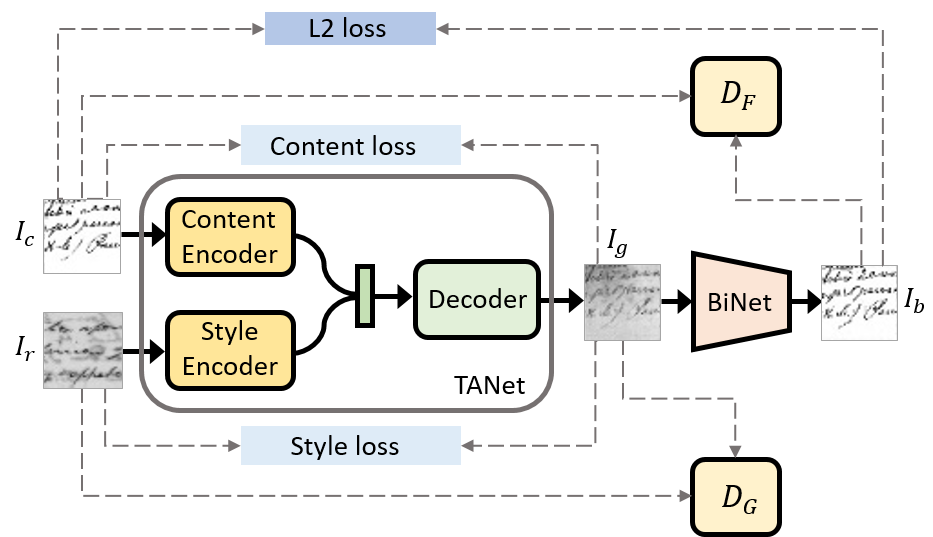}
  \caption{Illustration of the Proposed framework: It consists of two networks TANet and BiNet. TANet takes a clean document image $I_{c}$ and a degraded reference image $I_{r}$ and tries to generate an image $I_{g}$ with same textual content as $I_{c}$ retaining the noisy texture of $I_{r}$. Next, BiNet tries to get back the clean image by de-noising the generated one.}
  \label{figure:1}
\end{figure}

Early document binarization methods \cite{otsu1979threshold, sauvola2000adaptive, phansalkar2011adaptive, gatos2008improved} for binarization used various thresholding techniques, which include finding a single (or multiple) appropriate threshold(s) for classifying pixels in the image as belonging to foreground or background. 
Recently, few deep learning frameworks \cite{vo2018binarization, tensmeyer2017document} have also been applied for binarization of document images. The objective here is not to predict a threshold but to directly output a binary mask segregating the foreground text and the background noise. These deep learning based models require a considerable amount of paired training data. The publicly available binarization datasets are not sufficient to learn various possible noise distributions (i.e., artifacts, stains, ink spills, etc.) that may occur in real-life situations. 

In this paper, we intend to increase the utility of the available bounded datasets by proposing a novel adversarial learning technique. The basic idea is to generate a new set of augmented images of high perceptual quality that combine the semantic content of a clean binary image with the noisy appearance of a degraded document image. Thus, the low-level distribution of visual features in an image is modified while maintaining the semantic content. 
In this way, we can generate multiple degraded versions of same textual content with various noisy textures. For this purpose we propose a Texture Augmentation Network that superimposes the noisy appearance of the degraded document on the clean binary image. Then, the output image is passed through a Binarization Network called BiNet to get back the clean version of the document image. Both the networks are jointly trained in an adversarial manner. The goal is to generate harder adversarial training samples with lots of variations. On the other hand, BiNet tries to learn from the hard augmentations for better performance. By jointly training the two networks, we can enhance the adversarial robustness of our binarization model. The advantage of this technique is that the system can learn from unpaired images. It becomes very useful in case of ancient historical documents as it is difficult to get the corresponding binary images for these documents. If the system supports unpaired setting then the data collection process becomes easier. We can easily get large number of unpaired images with less effort by collecting document images and clean images independently from different sources. However, no other previous work on document binarization have tried to utilize the unpaired dataset. In this paper, the proposed system can be trained with unpaired data.  

The proposed framework is summarized in Figure \ref{pic1}.
The main contributions of our study are as follows: (1) To the best of our knowledge, our work is the first attempt to use a Generative adversarial model in document binarization problem. (2) We propose a texture augmentation network to augment image datasets, by generating adversarial examples online. (3) We employ adversarial learning technique driven by a general GAN objective where the GAN loss plays a complementary role in training TANet and BiNet jointly. Also, it is noteworthy that the method is able to learn from unpaired datasets.

\section{METHODOLOGY}
\label{sec:metho}

In this section, we present the details of our proposed binarization model. The model consists of two networks: Texture Augmentation Network (TANet) and 
Binarization Network (BiNet). Given a clean document image, TANet tries to obtain a noisy version of that image by transferring the noisy texture of a reference document image comprising of various degradations. On the other hand, the BiNet tries to binarize the newly generated noisy image.

\textbf{Texture Augmentation Network.} To combine the semantic content of a clean document image and the noisy texture of a degraded document image, the first step would be to separate the content and texture representation explicitly. For this purpose, we employ a Content encoder and a Style encoder. Given a clean image $I_{c}$ and reference noisy element $I_{r}$, the encoders learn to extract latent representations $R_{c}$ and $R_{r}$, by leveraging the conditional dependence of the content and texture images. Both encoders have the same configuration with eight convolutional blocks of filter size $5\times5$ and stride 2. Each convolutional layers are associated with Leaky ReLU and Batch Normalization. 

After extracting the content and texture representations, we perform simple concatenation to obtain a mixed representation. Then, it is passed through a Decoder network that maps the combined representations to an output image that has the same textual content as the clean image and the same texture element as the noisy input. The Decoder architecture is symmetrical to the Encoders with series of deconvolution-BatchNorm-LeakyReLU up-sampling blocks with tanh activation for the final output. 
The output and the clean image differ in appearance, but they have the same textual element. However, due to the down-sampling process in the content encoder, only a part of the input is stored, resulting in a significant information loss in each layer which can not be used to generate the output image. To deal with this, we adopt skip-connection between the layers of the content encoder and the decoder. We concatenate the feature-map of each down-sampling block in the content encoder with the corresponding feature map of the up-sampling block in the decoder.
We represent the TANet as $G$. 
 
\begin{equation}
 I_{g} = G(I_{c}, I_{r}; \theta_{G})
\end{equation}  

where, $I_{g}$ is the generated output and $\theta_{G}$ is the parameters of TANet. The image generated by the TANet should follow some constraints and objectives: (1) It should look real and can not be distinguished from the real-world noisy, degraded document images (2) It has similar texture appearance as the degraded reference document image $I_{r}$. (3) It has same textual content as the clean document image $I_{c}$.  To incorporate above constraints in the training process of TANet, we adopt the following loss functions for TANet.

\textit{Adversarial loss.} we use an adversarial objective to constrain the output to look similar to the reference document image. Assuming $I_{r}$ is sampled according to some data distribution $\mathcal{P}_{r}$ and $I_{c}$ is sampled from distribution $\mathcal{P}_{c}$, the loss is defined as 

\begin{multline}
L_{G}^{GAN}(G,D_{G}) = \E_{I_{r}\in \mathcal{P}_{r}}[logD_{G}(I_{r})]+\\
\E_{I_{c}\in \mathcal{P}_{c},I_{r}\in \mathcal{P}_{r} }[log(1-D_{G}(G(I_{c}, I_{r})))]
\end{multline}

where the discriminator $D_{G}$ tries to discriminate between the output image from the degraded reference image. 

\textit{Style loss.} the adversarial loss focuses on getting the overall structure but sometimes it is not enough to capture the fine details of the texture. We use an additional style loss $L^{S}$ to ensure the successful transfer of texture from the reference image to the clean document image. Following \cite{gatys2015texture,gatys2016image}, we have used the technique of matching the gram matrices. It captures the correlations between the different feature responses extracted from certain layers of a pre-trained VGG-19 network. Mathematically, gram matrix $\mathcal{G}_{ij}^{l} \in \mathcal{R}^{N_{l} \times N{l}}$ is the inner product between
the vectorised feature maps $i$ and $j$ in layer $l$:
\begin{equation}
\mathcal{G}_{ij}^{l} = \sum_{k}\mathcal{F}_{ik}^{l}\mathcal{F}_{jk}^{l}
\end{equation}
where, $N_{l}$ is the number of feature maps and $\mathcal{F}_{ik}^{l}$ is the activation of $i^{th}$ filter at position $k$ in layer $l$.  We use 5 layers of VGG-19 network ("conv1\_1", "conv2\_1", "conv3\_1", "conv4\_1", "conv5\_1") to define our style loss.

\textit{Content loss.} It is required that the generated images have the same textual content as the clean document image. To incorporate the same in our training process, we define a masked mean square loss function. The loss penalizes the differences between the pixels of the content image and output image in the text region only. It can be defined as 

\begin{equation}
L^{c}(G) = ||M \odot I_{c} - M \odot I_{g}||_{2}
\end{equation}

where, $M$ is a binary mask that has value 1 in the text region and 0 in the background. Thus, the total objective function to train TANet can be written as  

\begin{equation}
L^{TANet} = L_{G}^{GAN}(G,D_{G}) + \lambda_{s}L^{s}(G) +  \lambda_{c}L^{c}(G)
\end{equation}
where, $\lambda_{s}$ and $\lambda_{c}$ are the weights to balance the multiple objectives. 

\textbf{Binarization Network.} Given that we have generated the noisy version of a clean document image, our system tries to get back the clean binarized image from the generated one through another network called BiNet. The network employs an image-to-image translation framework consisting of a generator and a discriminator. The objective is to train a generator network $F$ that takes the output of TANet $I_g$ and obtains a binarized version of that image $I_b$. 
\begin{equation}
I_b = F(I_g; \theta_F)
\end{equation}
where, $\theta_F$ is the parameters of the network $F$. A discriminator network $D_F$ is used to determine how good the generator is in generating binarized images. We have used similar network architecture for the generator and the discriminator as mentioned in \cite{konwer2018staff}. During training, both the networks compete against each other in a min-max game. The training objective can be defined as 

\begin{multline}
L_{F}^{GAN}(F,D_{F}) = \E_{I_{c}\in \mathcal{P}_{c}}[logD_{F}(I_{c})]+\\
\E_{I_{g}\in \mathcal{P}_{g}}[log(1-D_{F}(F(I_{g})))]
\end{multline}

It is noted that the training, in this case, follows the "paired" setting. For each input image to the network $F$, there is corresponding ground truth image. Thus, we can employ full supervision on the predicted binarization results by leveraging the advantage of $L_2$ pixel loss along with the adversarial loss. 

\begin{equation}
L^{L2}(F) = ||I_c-I_b||_2
\end{equation}

The adversarial loss helps to obtain sharper output image by de-noising the noisy input whereas the $L_2$ loss helps to preserve the content. The final objective of BiNet is 

\begin{equation}
L^{BiNet} = L_{F}^{GAN}(F,D_{F}) + \lambda_{L2}L^{L2}(F) 
\end{equation}

where, $\lambda_{L2}$ is a weight parameter. In the next section, we will provide some salient details of the training process and discuss the appropriate weights.

\section{Experiments}
\label{sec:exp}

In this section we discuss about the datasets, training details, baseline methods and experimental results regarding the evaluation of our proposed binarization model. 

\textbf{Datasets.} For training and evaluating our model, we have used some publicly available document datasets. A total of 9 datasets are used in this work: DIBCO 2009 \cite{gatos2009icdar}, DIBCO 2011 \cite{pratikakis2011icdar}, DIBCO 2013 \cite{pratikakis2013icdar}, H-DIBCO 2010 \cite{pratikakis2010h}, H-DIBCO 2012 \cite{pratikakis2012icfhr}, H-DIBCO 2014 \cite{ntirogiannis2014icfhr2014}, Bickley diary \cite{deng2010binarizationshop}, PHIDB \cite{nafchi2013efficient}, and S-MS \cite{hedjam2015icdar} datasets. Out of these datasets, DIBCO 2013 dataset is selected for testing purposes. For the testing, the remaining datasets are used as a training set. At first, we convert the images from these datasets to patches of size $256\times256$. To increase the number of patches, we augment the training patches by rotating with an angle of 90, 180, or 270. A small part (10\%) of the obtained image patches is used as an evaluation set, and the rest of the images are used to train the model. In the training set, we have two set of images patches: degraded document set and as well as their binarized ground truths. The clean images are sampled from the binarized set, and reference images are sampled from the degraded document set in an unpaired manner.  

\begin{figure*}
\label{pic2}
  \includegraphics[width=1\linewidth]{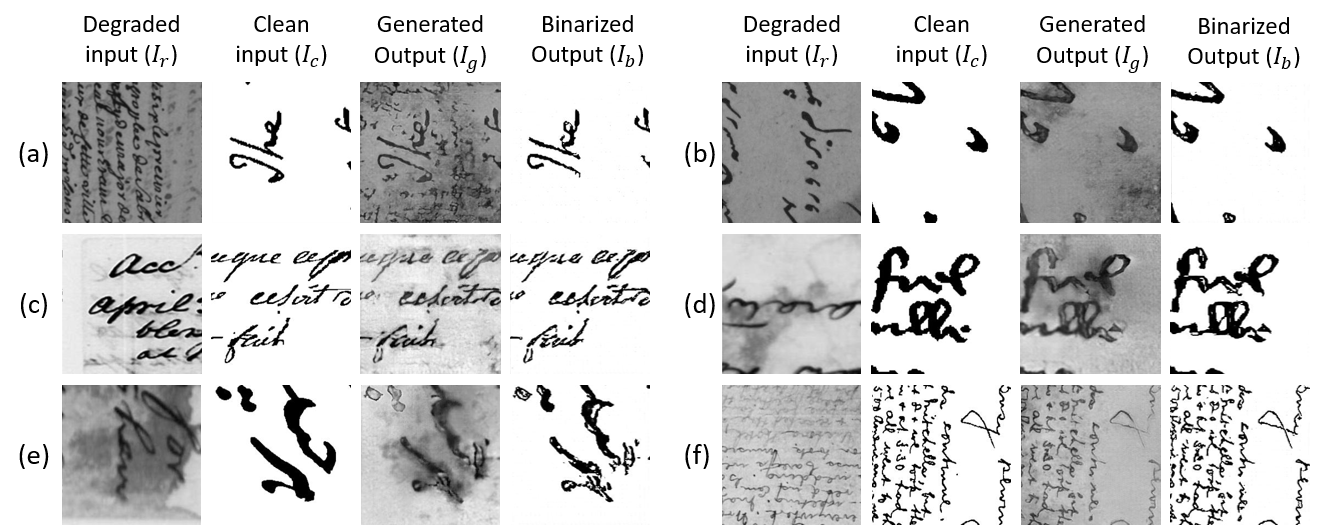}
  \caption{Qualitative evaluation of texture transfer process and binarizarion technique on the evaluation set.}
  \label{figure:1}
\end{figure*}

\textbf{Training.} To train the model, we follow a particular stage-wise training protocol. At first, TANet is trained for 10 epochs. After the training, it should be able to generate noisy version of the clean images. In the next stage, BiNet is trained using the generated noisy images for another 10 epochs. At last, TANet and BiNet are fine-tuned together for around 30 epochs. We note that during the couple training, TANet tends to generate more challenging adversarial samples that are relatively hard to be detected by BiNet. This training strategy enforces the model to learn a various type of degradations including noises and artifacts. Figure \ref{pic2} illustrates the texture transfer process qualitatively. At the time of testing, BiNet is used to get the binarized output of a given document image. Experiments are conducted on a server with 12 GB memory and single Nvidia Tesla K80 GPU. The model is implemented using TensorFlow library. Adam optimizer with learning rate 0.0001 is used to train the model. We take $\lambda_{s}$ = 0.5, $\lambda_{c}$ = 10 and $\lambda_{L2}$ = 100 throughout the experiments. We used the following metrics to quantitatively measure the performance of our proposed model with those of the state-of-the-art algorithms and some baselines: F-measure, pseudo-F-measure ($F_{ps}$), Distance reciprocal distortion metric (DRD), and the peak signal-to-noise ratio (PSNR) metrics \cite{pratikakis2010h}. 

\setlength\textfloatsep{3mm}
\textbf{Baselines:} We define following baselines.

\textit{U-Net:} It is a trivial encoder-decoder network with skip-connections \cite{ronneberger2015u}. We take its same architecture as our generator unit of BiNet. It is noted that the network is trained in a paired setting. For each input image, there is corresponding ground truth. L2 pixel loss is used to train the complete model.

\textit{Pix2pix:} It is an image-to-image translation framework inspired from \cite{isola2017image}. The network resembles the BiNet part of our system. It is trained using adversarial loss and L2 loss using paired data. 

\textit{CycleGAN:} We employ this baseline by using the concept of cycle-consistent image translation frameworks \cite{zhu2017unpaired}. The network utilizes unpaired data to train the model. 
\setlength\textfloatsep{3mm}
\begin{table}[]
\label{tab}
\centering

\begin{tabular}{l|c|c|c|c}
\hline
\textbf{Methods} & \textbf{F-measure} & $\mathbold{F_{ps}}$ & \textbf{DRD} & \textbf{PSNR} \\ \hline
BERN \cite{bernsen1986dynamic}             & 52.6               & 52.8              & 62.2         & 10.1          \\ \hline
Sauvola \cite{sauvola2000adaptive}        & 85.0               & 89.8              & 7.6          & 16.9          \\ \hline
Niblack \cite{niblack1986introduction}          & 72.8               & 72.2              & 24.9         & 13.6          \\ \hline
Gatos \cite{gatos2004adaptive}           & 83.4               & 87.0              & 9.5          & 17.1          \\ \hline
Otsu \cite{otsu1979threshold}            & 83.9               & 86.5              & 11.0         & 16.6          \\ \hline
Su \cite{su2013robust}              & 87.7               & 88.3              & 4.2          & 19.6          \\ \hline
Howe \cite{howe2013document}           & 91.3               & 91.7              & 3.2          & 21.3          \\ \hline
DSN \cite{vo2018binarization}              & 94.4               & 96.0              & 1.8          & 21.4          \\ \hline
U-Net \cite{ronneberger2015u}           & 89.6               & 91.8              & 5.9          & 17.6          \\ \hline
Pix2pix \cite{isola2017image}         & 94.8               & 97.0              & 2.7          & 20.8          \\ \hline
CycleGAN \cite{zhu2017unpaired}        & 66.8               & 70.1              & 17.6          & 12.5          \\ \hline
\textbf{Ours}    & \textbf{97.8}      & \textbf{98.7}     & \textbf{1.1} & \textbf{24.3} \\ \hline
\end{tabular}
\caption{Quantitative results on DIBCO 2013 dataset}
\end{table}

\begin{figure}
\label{pic3}
  \includegraphics[width=1\linewidth]{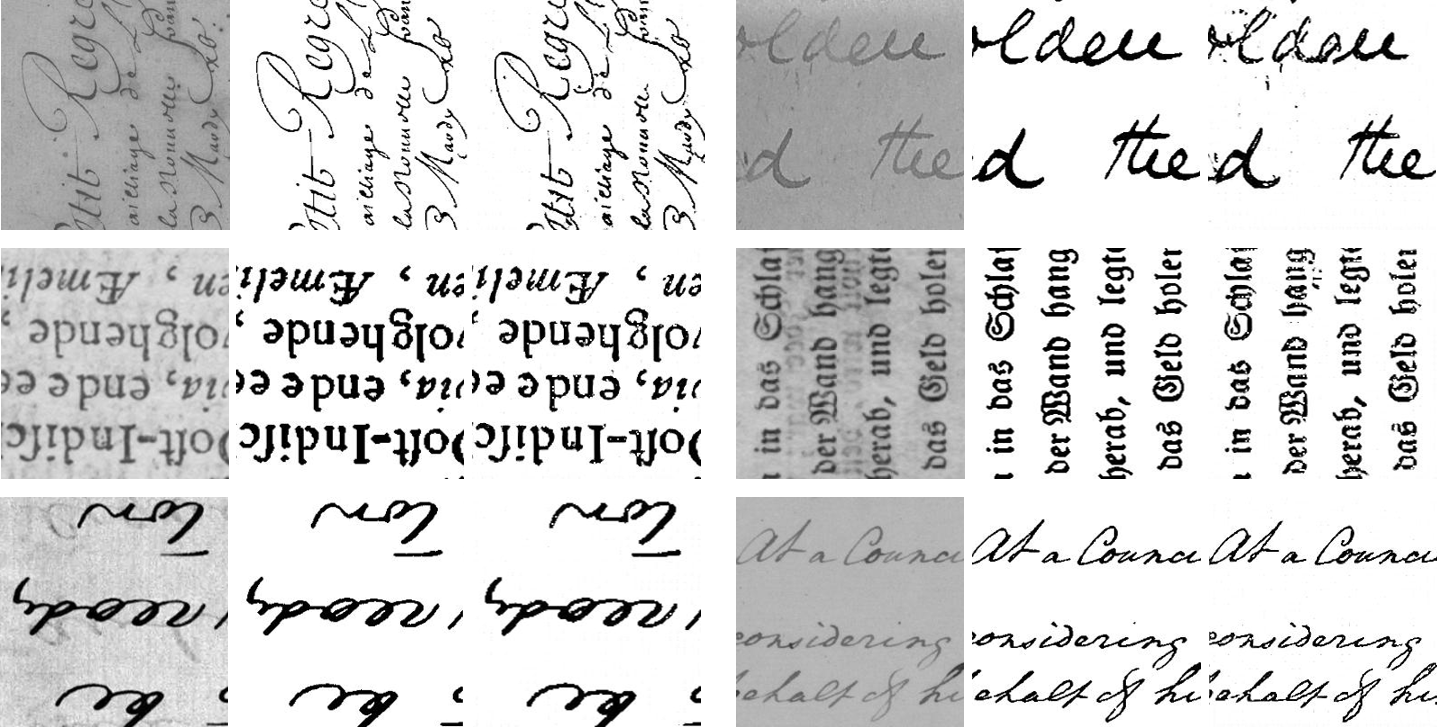}
  \caption{Binarizartion results using the trained BiNet model on test set. Images are in the order: input, predicted and ground truth from left to right for each sample result.}
\end{figure}
\setlength\textfloatsep{2mm}
\textbf{Results:}We compare our proposed system with the baseline methods and some state-of-the-art binarization algorithms in Table \ref{tab}. Some qualitative results are shown in Figure \ref{pic3}. From Table \ref{tab}, we can see that our proposed method delivers the best quality results regarding all the four evaluation metrics. Also, we have obtained a low DRD score which implies that our method is also superior regarding the visual distortion. U-Net and Pix2pix work moderately, but CycleGAN obtains poor results as compared to others. Similar to CycleGAN our method also utilizes unpaired data, but the main binarization network (BiNet) of our model learns from paired samples which are created internally in our system. Thus, we can impose full supervision in the BiNet part that helps to generate high-quality results. However, in CycleGAN method, there is no scope to impose the full supervision. 

\section{CONCLUSION}
In this paper, we re-visited the problem of document binarization by introducing a new adversarial learning technique that intends to increase the utility of the available bounded datasets. The noisy data augmentation is an integral part of our network that enforces the model to learn robust representation of various types of document degradations from unpaired data. Also, the experimental results suggest that our method is superior to existing state-of-the-art frameworks.  
\vfill\pagebreak

\bibliographystyle{IEEEbib}
\bibliography{strings,refs}

\end{document}